\documentclass[letterpaper]{article} 
\usepackage[preprint]{aaai2027}
\usepackage[hyphens]{url}  
\usepackage{graphicx} 
\usepackage{natbib}  
\usepackage{caption} 
\usepackage{algorithm}
\usepackage{algorithmic}

\usepackage{newfloat}
\usepackage{listings}
\usepackage{amssymb}  
\usepackage{amsmath}
\usepackage{multirow}
\usepackage[table]{xcolor}  
\DeclareCaptionStyle{ruled}{labelfont=normalfont,labelsep=colon,strut=off} 
\floatstyle{ruled}
\newfloat{listing}{tb}{lst}{}
\floatname{listing}{Listing}

\usepackage{booktabs}

\title{Compressed Video Aggregator: Content-Driven Module for\\ Efficient Micro-Video Recommendation}
\author{
    Yang Xiao\textsuperscript{\rm 1 5},
    Huiyuan Chen\textsuperscript{\rm 2},
    Kaiyuan Deng\textsuperscript{\rm 3},
    Chao Jiang\textsuperscript{\rm 4},\\
    Zinan Ling\textsuperscript{\rm 1},
    Ruimeng Ye\textsuperscript{\rm 1},
    Fei Wang\textsuperscript{\rm 5},
    Xiaolong Ma\textsuperscript{\rm 3},
    Bo Hui\textsuperscript{\rm 1}\corresponding
}
\affiliations{
    \textsuperscript{\rm 1}University of Tulsa 
    \textsuperscript{\rm 2}Amazon
    \textsuperscript{\rm 3}University of Arizona
    \textsuperscript{\rm 4}Visa
    \textsuperscript{\rm 5}Alchedata.AI
}

\begin{document}

\maketitle

\begin{abstract}
We propose \textbf{Compressed Video Aggregator} (CVA), a lightweight micro-video recommendation module that decouples video information from preference learning. CVA first summarizes frozen VFM frame embeddings into a semantic-consensus anchor through masked mean pooling, projects this anchor into a compact latent space, and refines the projected representation with residual self-attention and feedforward blocks before producing a single video embedding for the recommender. Due to the redundancy in the frame count of the original benchmark and its overly coarse sampling, we used titles to re-select key frames based on CLIP. Experiments on MicroLens and Short-Video show consistent gains with orders-of-magnitude reductions in training time and GPU memory, and re-selected frames can further enhance the performance of all methods, including CVA. Furthermore, we also discussed the impact of several scenarios involving erroneous titles on our method. 
\end{abstract}


\begin{links}
    \link{Code}{https://github.com/NKUShaw/LiteFrame}
\end{links}
\section{Introduction}
Micro videos, also known as short-form videos, have become a popular and widespread form of entertainment in recent years. These videos typically range in length from a few seconds to several minutes and exist on various platforms, including social media, video-sharing websites, and mobile applications. The number of micro-video users worldwide has exceeded several billion worldwide, especially with 1.05 billion in China alone~\cite{CNNIC2025}. The surge in popularity of micro-videos has motivated advanced research in micro-video recommendation. To improve user experience and user engagement, various personalized micro-video recommendation methods~\cite{gong2022real,zheng2022dvr,yu2022improving,han2016dancelets,liu2021concept,gao2022kuairec,yuan2022tenrec,liu2019user,jiang2020aspect,lei2021semi,DBLP:conf/sigir/HeL0GMLZZ25,DBLP:conf/kdd/XuPLX25} have been proposed.

Existing video recommendation methods face two critical limitations: (i) The majority of existing works rely heavily on user-video interaction signals such as clicks, watch time, and dwell time, or video-level attributes such as titles, covers, likes, and views~\cite{zheng2022dvr,yu2022improving, gong2022real}. Without modeling the video content itself, these methods fail to directly capture the semantic content or temporal structure of videos and may generalize poorly beyond observed interaction patterns. (ii) Directly leveraging video content is computationally prohibitive.
Modern short-video platforms contain millions of videos and massive volumes of user activity. Applying sophisticated video encoders~\cite{ni2025content}, Vision-Language Models (VLMs)~\cite{de2025describe}, or Video Multimodal Large Language Models~\cite{DBLP:conf/iclr/0002H0ZLHZWCW25} to representation learning from raw videos introduces substantial overhead: a) Extremely long training time: temporal modeling requires large-scale multi-GPU training; b) High GPU memory consumption: video models process multiple frames simultaneously, far exceeding image-level resource demands; c) Slow inference latency: online recommendation cannot afford per-video feature extraction with heavy encoders. These constraints make large-scale video-aware recommendation systems difficult to deploy in practice, despite the clear need for content-based modeling in short-video platforms.

In this paper, we propose a lightweight method for training content-driven micro-video recommendation models online, with title guidance for enhanced offline frame sampling. Our work consists of two key components: \textbf{semantic resampling} and \textbf{video compression}.

\textbf{Semantic Resampling}. In most video tasks, models do not take the full video as input. Instead, the standard strategy is to reduce temporal redundancy through uniform sampling~\cite{shang2025large,liu2022video,feichtenhofer2019slowfast,bertasius2021space} or random sampling~\cite{tong2022videomae}, allowing the model to process only a compact but informative subset of frames. Among existing video recommendation datasets that provide the raw video content itself, Short-Video~\cite{shang2025large} adopts uniform sampling of 8 frames, while MicroLens~\cite{ni2025content} extracts five consecutive frames at the temporal midpoint of each video. However, the semantic information within a video is not necessarily distributed in a regular or predictable manner, and an inappropriate sampling strategy may yield frames that fail to represent the video's overall meaning. Following the intuition of uniform sampling, we first partition each video into 100 clips and extract 100 uniformly distributed frames. We then employ a CLIP model~\cite{radford2021learning} to select the five to eight frames most semantically aligned with the video title. Through video description/caption quality analysis (Table~\ref{tab:desc_evaluation}) and evaluation on downstream video recommendation tasks (Table~\ref{tab:experiments}), we demonstrate that our sampling strategy is more effective than both uniform sampling and random sampling.

\begin{figure}[t]
    \centering
    \includegraphics[width=1.0\linewidth]{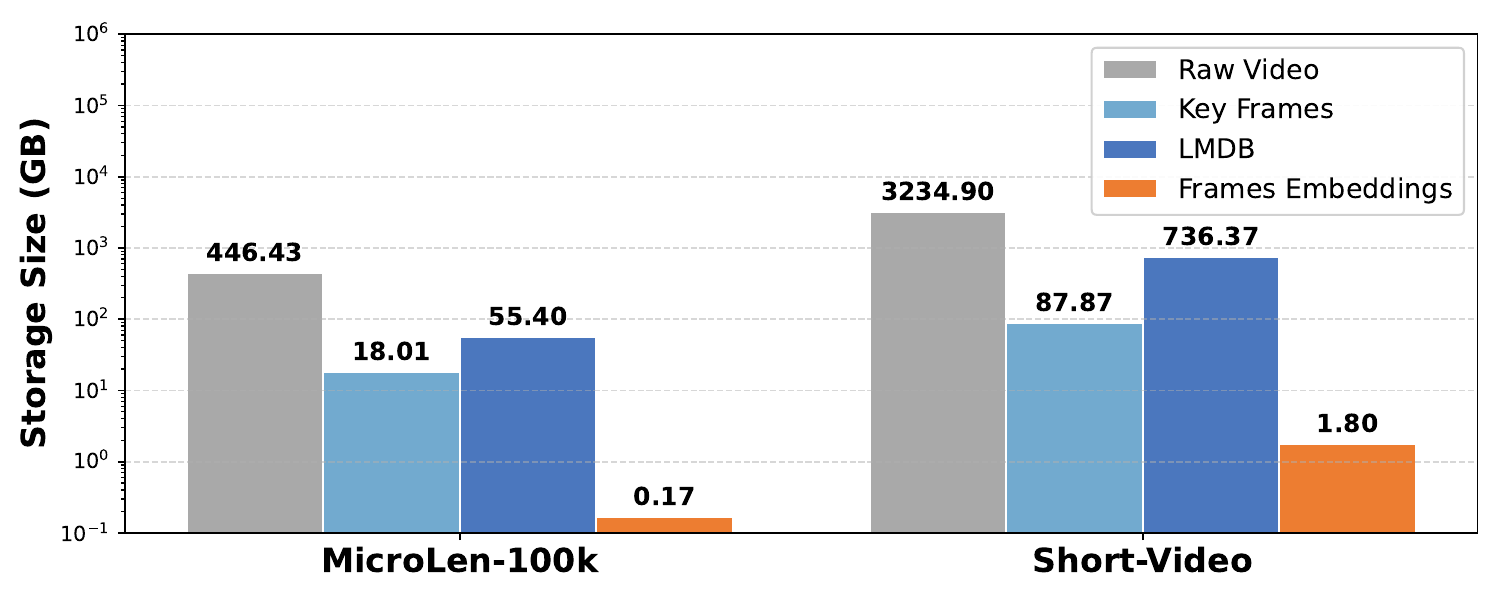}
    \caption{Data size comparison.}
    \label{fig:data_size_comparsion}
\end{figure}

\textbf{Video Compression.}
While resampling reduces temporal redundancy, it does not address the computational cost of high-dimensional visual features.
Standard visual foundation models (e.g., DINOv3, CLIP) produce dense features that remain too costly for sequential recommendation, even with few sampled frames.
Previous studies~\cite{ni2025content} showed that directly using frozen visual backbone models yields poor recommendation performance. Although end-to-end fine-tuning substantially improves accuracy, it incurs prohibitive training latency and GPU memory consumption. Our experimental results confirm the same trend (Table~\ref{tab:experiments}).
To bridge this gap, we introduce an efficient \textbf{Compressed Video Aggregator (CVA)}. CVA first mean-pools the cached frame embeddings into a video-level semantic anchor, projects this anchor into a compact latent space, and then applies lightweight latent refinement to produce a \textbf{Compact Video Embedding (CVE)}.
Here, we show a comparison example of the compressed size (See in Figure~\ref{fig:data_size_comparsion}).
This module acts as a semantic projector that compresses the dense visual features of the selected key frames into a single input-conditioned latent representation. Because frame aggregation precedes latent refinement, CVA models video-level semantic consensus rather than pairwise temporal interactions between frames. This design retains the semantic knowledge of large visual models while providing inference speed and memory efficiency comparable to traditional ID-based recommendation systems.

The main contributions of this work are:

\begin{itemize}
    \setlength{\topsep}{0pt}
    \setlength{\partopsep}{0pt}
    \setlength{\itemsep}{0pt}
    \setlength{\parskip}{0pt}
    \setlength{\parsep}{0pt}

    \item \textbf{Lightweight Video Embeddings.}
    We extract compact and reusable frame embeddings using a frozen Visual Foundation Model (VFM), significantly reducing raw video redundancy for large-scale recommendation.

    \item \textbf{Efficient Video Recommendation.}
    We propose a video recommendation framework combining semantic resampling and a lightweight module \textbf{CVA}, enabling efficient video modeling with optional title-guided semantic enhancement while minimizing reliance on user or item metadata.

    \item \textbf{Strong Performance--Efficiency Trade-off.}
    Experiments on MicroLens and Short-Video show consistent gains in Hit and nDCG, while substantially reducing training time and GPU memory usage.
\end{itemize}

\begin{figure*}[t]
    \centering    \includegraphics[width=1.0\linewidth]{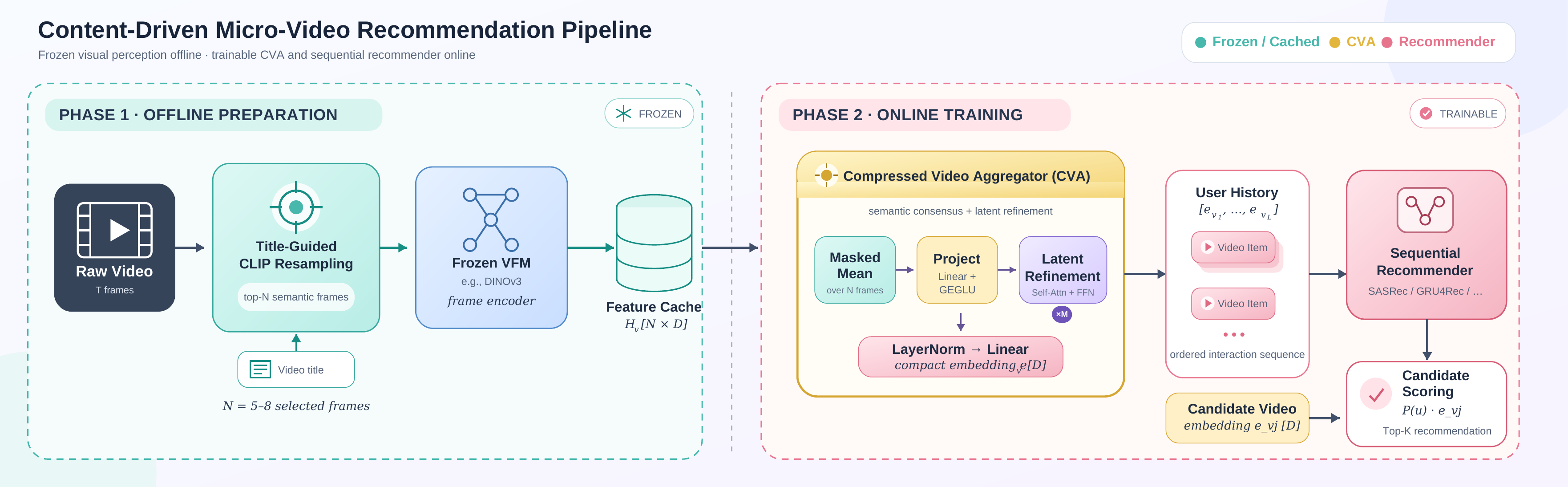}
    \caption{The overall pipeline of our proposed method. The framework is decoupled into two distinct phases to maximize efficiency: (1) Offline Preprocessing, where raw videos are resampled into key frames by the frozen CLIP and encoded by a frozen visual foundation model (e.g., DINOv3) into cached feature maps; and (2) Online Training, where the Compressed Video Aggregator (CVA) mean-pools the frame embeddings into an input-conditioned semantic anchor, refines the projected anchor in latent space, and outputs a single compact video embedding for the user sequence encoder.}
    \label{fig:pipeline}
\end{figure*}

\section{Related Work}
\noindent\textbf{Video Recommendation.} Early recommenders such as YoutubeDNN~\cite{covington2016deep}, SASRec~\cite{kang2018self}, and GRU4Rec~\cite{hidasi2015session} mainly model preferences from user--item interaction histories (e.g., ID sequences and watch time). While effective for sequential modeling, they struggle with cold-start items. Multimodal recommenders alleviate this by incorporating side information (e.g., text, cover images, audio, and user/item attributes)~\cite{wei2019mmgcn,yuan2023go,deldjoo2016content,he2016ups}, but typically treat content as auxiliary features while still relying on IDs~\cite{yuan2023go,gao2022kuairand,gao2022kuairec,Sun2023KuaiSAR}.

For micro-video platforms, raw frames carry the key visual and temporal signals that static metadata cannot capture, yet video-only recommendation remains underexplored. Early work~\cite{lee2017large} showed the feasibility of video-only cold-start modeling by learning compact video embeddings from visual features (Inception-v3~\cite{szegedy2016rethinking}). More recently, \textbf{MicroLens}~\cite{ni2025content} argues for training video-only recommenders directly on raw frames. In parallel, the \textbf{Short-Video} benchmark~\cite{shang2025large} provides rich multimodal signals, but lacks a systematic exploration of video-only recommendation. Table~\ref{tab:dataset_overview} compares the available information in these datasets.

\begin{table}[t]
    \centering
    \small
    \setlength{\tabcolsep}{0.5mm} 
        \begin{tabular}{lcccc}
            \toprule
            Dataset & U. Behav. & U. Attr. & V. & V. Attr. \\
            \midrule
            KuaiRec~\cite{gao2022kuairec}
                & \checkmark & \checkmark & $\times$ & \checkmark \\
            KuaiRand~\cite{gao2022kuairand}
                & \checkmark & \checkmark & $\times$ & \checkmark \\
            KuaiSAR~\cite{Sun2023KuaiSAR}
                & \checkmark & \checkmark & $\times$ & \checkmark \\
            TikTok~\cite{tiktok_10m_2025}
                & \checkmark & \checkmark & $\times$ & \checkmark \\
            Short-Video~\cite{shang2025large}
                & \checkmark & \checkmark & \checkmark & \checkmark \\
            MicroLens~\cite{ni2025content}
                & \checkmark & $\times$ & \checkmark & \checkmark \\
            \bottomrule
        \end{tabular}%
    \caption{Comparison of micro-video datasets. U. refers to User, V. refers to Video, Behav. refers to Behavior, and Attr. refers to Attribute.}
    \vspace{-5mm}
    \label{tab:dataset_overview}
\end{table}
\noindent\textbf{Visual Embedding Compression.} Bridging the gap between frozen visual encoders and downstream tasks requires an effective projection module. In the realm of Multimodal Large Language Models (MLLMs), two primary paradigms exist: (1) MLP-based projection (e.g., LLaVA~\cite{liu2023visual}), which maps visual tokens linearly. While simple, it preserves the full sequence length, leading to high computational costs when applied to multi-frame videos. (2) Cross-attention-based projection, represented by Perceiver~\cite{alayrac2022flamingo,li2025otter,xiao2025optimal} and Q-Former~\cite{li2023blip,dai2023instructblip}, extracts and compresses visual information into a fixed number of tokens through cross-attention. In video recommendation, however, existing methods commonly rely on aggregation techniques such as mean pooling or LSTM/GRU. CVA follows a different design from cross-attention projectors: it first forms a semantic-consensus anchor through mean pooling and then refines this compact, input-conditioned representation in latent space.

\section{Method}
As illustrated in Figure~\ref{fig:pipeline}, our method decouples the complex visual perception from the recommendation learning process into two phases: (1) \textbf{Offline Preprocessing}, where we leverage a frozen visual foundation model to encode resampled key frames into cached features; and (2) \textbf{Online Training}, where a lightweight \textbf{Compressed Video Aggregator (CVA)} constructs a mean-pooled semantic anchor and refines it in a compact latent space for user preference modeling.

\subsection{Preliminary}
Let $\mathcal{U}$ and $\mathcal{V}$ denote the set of users and micro-videos, respectively. For each user $u \in \mathcal{U}$, we have a historical interaction sequence $S_u = \{v_1, v_2, \dots, v_t\}$, where $v_i \in \mathcal{V}$ represents the $i$-th video interacted with by the user, and $t$ is the sequence length.
Unlike traditional ID-based methods where each item $v$ is represented by a randomly initialized embedding table, we aim to learn the item representation directly from its raw visual content.
Each video $v$ consists of a sequence of raw frames $\mathcal{F}_v = \{f_1, f_2, \dots, f_T\}$. Our goal is to map $\mathcal{F}_v$ to a compact embedding $\mathbf{e}_v \in \mathbb{R}^d$ and subsequently learn a recommendation model to predict the next video $v_{t+1}$ that the user is likely to watch.

\subsection{Phase I: Offline Preprocessing (One-Shot)}
Directly processing all $T$ frames of a raw video during online training is computationally prohibitive due to the high dimensionality of visual data. To address this, we perform offline preprocessing to extract dense visual features.

\subsubsection{Semantic Resampling}
Video content often contains temporal redundancy. Instead of standard uniform or random sampling, we employ a semantic-based resampling strategy. We partition the raw video into 100 clips or segments, and utilize a VLM (CLIP\cite{radford2021learning}) to select the top-$N$ most informative frames based on semantic relevance to the video attributes (Title). There are 30 frames per second, and all videos have more than one hundred frames. We extract the middle frame in the 100 segments first. Then we compute the similarity between frames and the title via the CLIP model. Select the top-N frames in CLIP similarity. And finally, sort by timestamp. 
Let the selected key frames be denoted as $\hat{\mathcal{F}}_v = \{\hat{f}_1, \dots, \hat{f}_N\}$, where $N$ is significantly smaller than the original frame count $T$ (e.g., $N=5 \text{ in MicroLens} \text{ or } 8 \text{ in Short-Video}$).

\begin{figure*}[t]
    \centering
\includegraphics[width=1.0\linewidth]{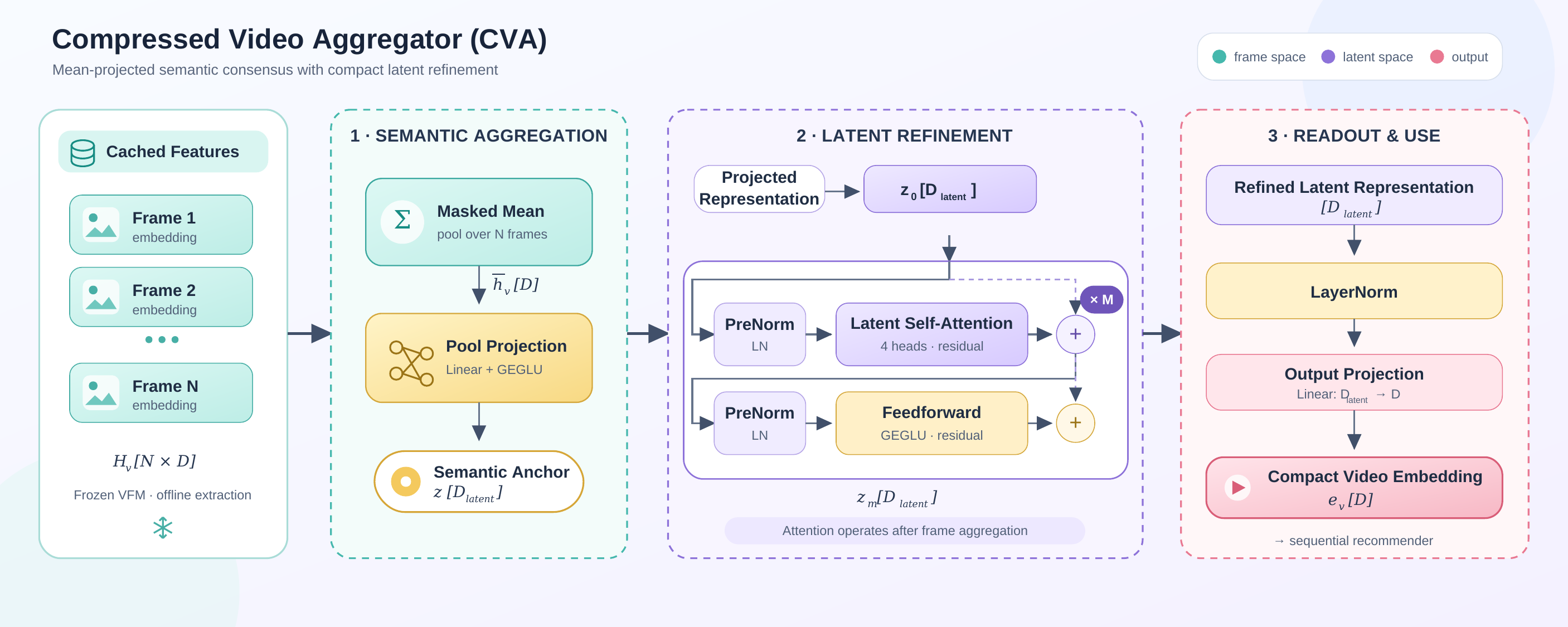}
    \caption{Compressed Video Aggregator (CVA) summarizes $N$ frame embeddings into a semantic-consensus anchor, projects the anchor into a compact latent representation, refines it with $M$ residual self-attention blocks, and decodes it into a compact video embedding.}
    \label{fig:CVA}
\end{figure*}

\begin{table*}[t]
    \centering
    \begin{tabular}{lcccc}
        \toprule
        \midrule
        Method & Semantic Sim. ($\uparrow$) & Perplexity ($\downarrow$) & Diversity ($\uparrow$) & Avg. Length ($\uparrow$) \\
        \midrule
        Baseline (Mid-5) & \textbf{0.5781} & 168.34 & 0.7336 & 103.80 \\
        Ours (Resample) & 0.5518 & \textbf{65.83} & \textbf{0.9496} & \textbf{165.13} \\
        \midrule
        \bottomrule
    \end{tabular}
    \caption{Quantitative evaluation of generated descriptions. Semantic Sim. denotes the BGE-M3 similarity with the video title. Perplexity measures text fluency, while Diversity (Distinct-2) and Avg. Length measures information richness. Our method achieves significantly better text quality and informativeness with comparable semantic alignment.}
    \label{tab:desc_evaluation}
\end{table*}

\subsubsection{Frozen Feature Extraction}
We employ pre-trained Visual Foundation Models (VFM), such as DINOv3, as the visual encoder $\mathbf{VFM}(\cdot)$. This encoder is \textbf{frozen} to avoid the massive computational cost of back-propagation through the vision backbone. For each selected frame $\hat{f}_i$, we extract its feature vector: $\mathbf{h}_i = \mathbf{VFM}(\hat{f}_i) \in \mathbb{R}^D$. Consequently, each video $v$ is represented as a dense feature matrix $\mathbf{H}_v = [\mathbf{h}_1, \dots, \mathbf{h}_N] \in \mathbb{R}^{N \times D}$. These features are saved to serve as the input for the online training phase. Results can be seen in Table~\ref{tab:vfm_statistics}.

The cache is constructed once per video and is shared by every training epoch, user sequence, candidate-scoring operation, and downstream user encoder. Neither semantic resampling nor the VFM participates in the online computation graph. Therefore, changing or retraining the recommender does not require repeated visual perception: online optimization loads only $\mathbf{H}_v$ and updates CVA together with the sequential user encoder. This separation converts the dominant visual-encoding cost from a repeated training expense into a one-shot preprocessing cost.

\begin{table}[t]
    \centering
    \small
    \begin{tabular}{lccc}
            \toprule
            VFM & Time (s) & Size (MB) & Dim. \\
            \midrule
            DINOv3 (ViT-S/16) & 189 & 172.1 & 384 \\
            DINOv2 (ViT-S/14) & 159 & 153.6 & 384 \\
            CLIP (ViT-B/32)   & 101 & 222.5 & 512 \\
            ViT-B/32          & 206 & 298.2 & 768 \\
            \bottomrule
        \end{tabular}%
    \caption{VFM statistics on MicroLens.
    Offline preprocessing is performed only once, and its cost remains
    substantially lower than that of end-to-end visual baselines.}
    \label{tab:vfm_statistics}
\end{table}

\subsection{Phase II: Online Training}
Although offline preprocessing reduces the temporal dimension to $N$, the cached embedding matrix $\mathbf{H}_v\in\mathbb{R}^{N\times D}$ is not directly compatible with sequential user modeling, which requires a compact 1D vector representation per item. To bridge this gap, we introduce the \textbf{Compressed Video Aggregator (CVA)} (Figure~\ref{fig:CVA}), a learnable module that separates frame-set aggregation from latent-space refinement. Its dimensional flow is $\mathbf{H}_v\in\mathbb{R}^{N\times D}\rightarrow\mathbf{z}\in\mathbb{R}^{D_{latent}}\rightarrow\mathbf{z}_M\in\mathbb{R}^{D_{latent}}\rightarrow\mathbf{e}_v\in\mathbb{R}^{D}$. Frame embeddings are aggregated once into a semantic anchor; the subsequent blocks operate only on its compact latent representation.
\subsubsection{CVA Encoder}
The encoder provides a streamlined alternative to cross-attention-based projectors such as Perceiver IO~\cite{jaegle2021perceiver}. The variable-length frame sequence $\mathbf{H}_v$ is first compressed into a single semantic-consensus vector by masked mean pooling along the frame dimension. A linear projection followed by GEGLU~\cite{hendrycks2016gaussian} maps this vector from $D$ to the latent dimension $D_{latent}$:
\begin{equation}
    \mathbf{z} = \operatorname{GEGLU}(\mathbf{W}_{p}\operatorname{MaskedMean}(\mathbf{H}_v)) \in \mathbb{R}^{D_{latent}}.
\end{equation}

\subsubsection{CVA Latent Refinement}
The projected anchor $\mathbf{z}_0=\mathbf{z}$ is passed through $M$ residual Transformer blocks, where $M$ is controlled by \texttt{CVA\_M}. Each block applies pre-normalized self-attention followed by a pre-normalized feedforward network:
\begin{equation}
    \mathbf{z}_{m} = \text{Block}(\mathbf{z}_{m-1}) \quad \text{for } m = 1, \dots, M
\end{equation}
Because the frame dimension has already been collapsed by masked mean pooling, these blocks refine an input-conditioned latent representation; they do not directly perform temporal attention or pairwise interaction between individual frames. The shape of $\mathbf{z}_M$ remains $\mathbb{R}^{D_{latent}}$.

\subsubsection{CVA Decoder}

The decoder applies Layer Normalization to the refined representation:

\begin{equation}
     \mathbf{e}' = \operatorname{LN}(\mathbf{z}_M) \in \mathbb{R}^{D_{latent}}.
\end{equation}

The normalized latent representation is passed through a final linear layer $g$ to match the embedding dimension required by the sequential user encoder (e.g., SASRec):

\begin{equation}
    \mathbf{e}_v  = g(\mathbf{e}') \in \mathbb{R}^{D}.
\end{equation}

This $\mathbf{e}_v$ is the final compact video embedding that serves as the input item representation for the sequential user encoder, completing the task of distilling complex, variable-length visual input into a compact, fixed-size vector.

For one video, masked aggregation costs $O(ND)$, the input and output projections cost $O(DD_{latent})$, and $M$ latent-refinement blocks cost $O(MD_{latent}^{2})$. The resulting online complexity is therefore
\begin{equation}
O\!\left(ND + DD_{latent} + MD_{latent}^{2}\right),
\end{equation}
with activation memory $O(ND+MD_{latent})$. 

In contrast, Perceiver-style projectors repeatedly couple frame tokens and latent queries through cross-attention. CVA removes that repeated frame-to-latent path: its dependence on the number of cached frames occurs only in the initial masked mean. The same property makes CVA permutation-invariant over the selected frame set. Its self-attention blocks refine the pooled, input-conditioned representation and should not be interpreted as temporal attention between individual frames.

\subsubsection{User Encoding}
Consider a user's historical sequence of watched-video embeddings, $\mathbf{S}(u)=[\mathbf{e}_{v_1}, \dots, \mathbf{e}_{v_T}]$, where $u$ denotes the user and $T$ is the sequence length. We apply a standard sequential user encoder (e.g., SASRec) to model the user's temporal preferences: $\mathbf{P}(u) = \mathbf{UserEncoder}(\mathbf{S}(u))$.

For a user $u$ at time step $t$, the prediction score for any candidate video $v_j \in \mathcal{V}_{B}$ is calculated via the dot product: $r_{u,t,v_j} = \mathbf{P}(u) \cdot \mathbf{e}_{v_j}$.

The model is trained by minimizing the Cross-Entropy loss over the batch:
\begin{equation}
    \mathcal{L} = -\sum_{(u, t) \in \mathcal{B}} \log \frac{\exp(r_{u, t, v_t})}{\exp(r_{u, t, v_t}) + \sum_{v_j \in \mathcal{N}_{u}} \exp(r_{u, t, v_j})}
\end{equation}

where $v_t$ is the ground-truth video and
$\mathcal{N}_{u}=\mathcal{V}_{B}\setminus(\{v_t\}\cup\mathbf{S}(u))$
denotes the set of negative samples in the batch after excluding the ground-truth item and the user's history.

This allows our model to learn distinct video representations by contrasting the ground-truth video against other videos in the same batch.

\section{Experiment}
\begin{table*}[t]
  \begin{center}
    \begin{small}
      \begin{sc}
        \begin{tabular}{lccccccc}
            \toprule
            \midrule
            \multicolumn{8}{c}{\textbf{MicroLens}} \\
            \midrule
            User Encoder  & Visual Encoder & HR@10 & NDCG@10 & HR@20 & NDCG@20 & Time(s) & Mem. (GB)\\
            \midrule
            SASRec& slowfast-50   & 9.179& 4.988& 13.072& 5.968& 46513&127.68\\
            SASRec& video-mae    & 8.373& 4.456& 12.500& 5.495& 31571&80.75\\
            SASRec& r3d18    & 8.459& 4.475& 12.565& 5.507& 24864&81.83\\
            SASRec& r3d50    & 5.047& 2.537& 8.145& 3.316& 25627&143.85\\
            SASRec& c2d50    & 8.889& 4.765& 12.924& 5.780& 24074&108.44\\
            SASRec& Matched MLP & 10.185& \textbf{5.670}& 14.376& 6.724& \textbf{1359}&\textbf{0.50}\\
            SASRec& Perceiver IO & 10.009& 5.420& 14.331& 6.510& 2366&6.13\\
            SASRec& CVA (M = 4) & \textbf{10.281}& 5.647& \textbf{14.723}& \textbf{6.767}& 1395&0.51\\
            \midrule
            GRU4Rec& slowfast-50   & 9.294& 5.021& 13.576& 6.097& 31044&127.59\\
            GRU4Rec& video-mae   & 8.221& 4.406& 12.145& 5.392& 39588&80.65\\
            GRU4Rec& Matched MLP & 9.684& 5.272& 13.960& 6.347& \textbf{1303}& \textbf{0.43}\\
            GRU4Rec& Perceiver IO & \textbf{9.821}& \textbf{5.355}& \textbf{14.240}& \textbf{6.467}& 2262&6.10\\
            GRU4Rec& CVA (M=4) & 9.790& 5.324& 13.990& 6.381& 1350&0.44\\
            \midrule
            Qwen3-VL-2B-LoRA& CVA (M = 4) & 8.373& 4.382& 12.414& 5.399& 4747&5.63\\
            Qwen3-VL-2B-Full FT& CVA (M = 4) & 8.403& 4.444& 12.478& 5.469& \textbf{2515}&11.30\\
            InternVL3-2B-LoRA & CVA (M = 4) & 8.505 & 4.528 & 12.587 & 5.554 & 3511 & \textbf{5.17}\\
            InternVL3-2B-Full FT & CVA (M = 4) & \textbf{8.611} & \textbf{4.534} & \textbf{12.709} & \textbf{5.565} & 2547 & 10.36\\
            \midrule
            \bottomrule
            
        \end{tabular}
      \end{sc}
    \end{small}
  \end{center}
  \caption{Main results on MicroLens-100K. Video-backbone baselines fine-tune the benchmark-recommended blocks. For both SASRec and GRU4Rec, the parameter-matched MLP, Perceiver IO, and CVA use identical resampled DINOv3 features, physical batch size 240, and seed 42. Their time and memory are measured under this controlled protocol. For the LLM-based user encoders, test results are evaluated using the checkpoint selected by validation-based early stopping.}
  \label{tab:experiments}
\end{table*}

\subsection{Experiment Setups}
We follow the MicroLens Benchmark~\cite{ni2025content} hyperparameters unless a controlled protocol is stated explicitly. Within each comparison, all methods use the same user-encoder and optimization settings. For Short-Video~\cite{shang2025large}, this dataset does not provide user sequences. We preprocessed it into a format similar to MicroLens based on the metadata. We list the detailed preprocessing steps in the appendix~\ref{sec: Data Preprocessing for the Short-Video Dataset}. 

We report \textbf{Hit@N} (whether the ground-truth item appears in the top-$N$ list; higher is better) with $N\!=\!10$ and $20$, and \textbf{nDCG}, which further accounts for ranking positions by giving higher weight to top-ranked hits. We also include \textbf{training time} and peak \textbf{GPU memory}. Unless otherwise stated, the efficiency results follow the original benchmark runs. The six controlled aggregator rows in Table~\ref{tab:experiments} and the three-seed SASRec comparison below use a physical batch size of 240 for every method.

For the primary SASRec comparison among CVA, Perceiver IO, and the parameter-matched FFN control, all methods use the same resampled DINOv3 frame features, physical batch size 240, refinement depth $M=4$, and random seeds 1, 42, and 1000. We report the mean and sample standard deviation over these three runs. The remaining large-scale backbone and user-encoder experiments use a single run because of their substantially higher computational cost. All reported test results are evaluated from the checkpoint selected by validation performance.

\textbf{Parameter-matched FFN control.}
To distinguish the effect of attention-based refinement from that of additional trainable parameters, we replace every latent self-attention sublayer in CVA with a two-layer token-wise FFN while leaving the semantic pooling, projection, residual paths, GEGLU feedforward sublayers, decoder, optimizer, and training protocol unchanged. For latent width $d$, the replacement uses bias-free $d\!\rightarrow\!2d$ and biased $2d\!\rightarrow\!d$ linear layers, yielding $4d^2+d$ parameters, exactly matching the query, key/value, and output projections of the corresponding attention sublayer.

\begin{table*}[t]
  \centering
  \small
  \begin{tabular}{lccccc}
    \toprule
    Visual Aggregator & \# of Param. & HR@10 & NDCG@10 & HR@20 & NDCG@20\\
    \midrule
    Perceiver IO & 5,321,344 & $10.119 \pm 0.109$ & $5.508 \pm 0.079$ & $14.524 \pm 0.167$ & $6.617 \pm 0.093$\\
    Parameter-matched MLP & 4,504,960 & $10.012 \pm 0.444$ & $5.509 \pm 0.313$ & $14.311 \pm 0.567$ & $6.590 \pm 0.339$\\
    CVA (M = 4) & 4,504,960 & $\mathbf{10.259 \pm 0.183}$ & $\mathbf{5.664 \pm 0.124}$ & $\mathbf{14.648 \pm 0.182}$ & $\mathbf{6.771 \pm 0.120}$\\
    \bottomrule
  \end{tabular}
  \caption{Controlled SASRec comparison on MicroLens-100K. All methods use identical resampled frame features, physical batch size 240, and seeds \{1, 42, 1000\}; metrics are reported as mean $\pm$ sample standard deviation. The MLP is exactly parameter-matched to CVA, while Perceiver IO is an architectural baseline.}
  \label{tab:controlled_aggregators}
\end{table*}

CVA obtains the highest mean on all four ranking metrics. Compared with Perceiver IO, CVA improves HR@10/NDCG@10 by $+0.139/+0.156$ and HR@20/NDCG@20 by $+0.124/+0.154$. Against the exactly parameter-matched MLP, the corresponding gains are $+0.246/+0.155$ and $+0.337/+0.181$. The MLP also exhibits substantially larger variation across seeds, including standard deviations of $0.444$ versus $0.183$ for HR@10 and $0.567$ versus $0.182$ for HR@20. These results support that attention-based latent refinement improves both average recommendation quality and robustness beyond merely adding an equal number of parameters.

\subsection{Overall Performance Comparison}
\begin{figure}[t]
    \centering
\includegraphics[width=0.85\linewidth]{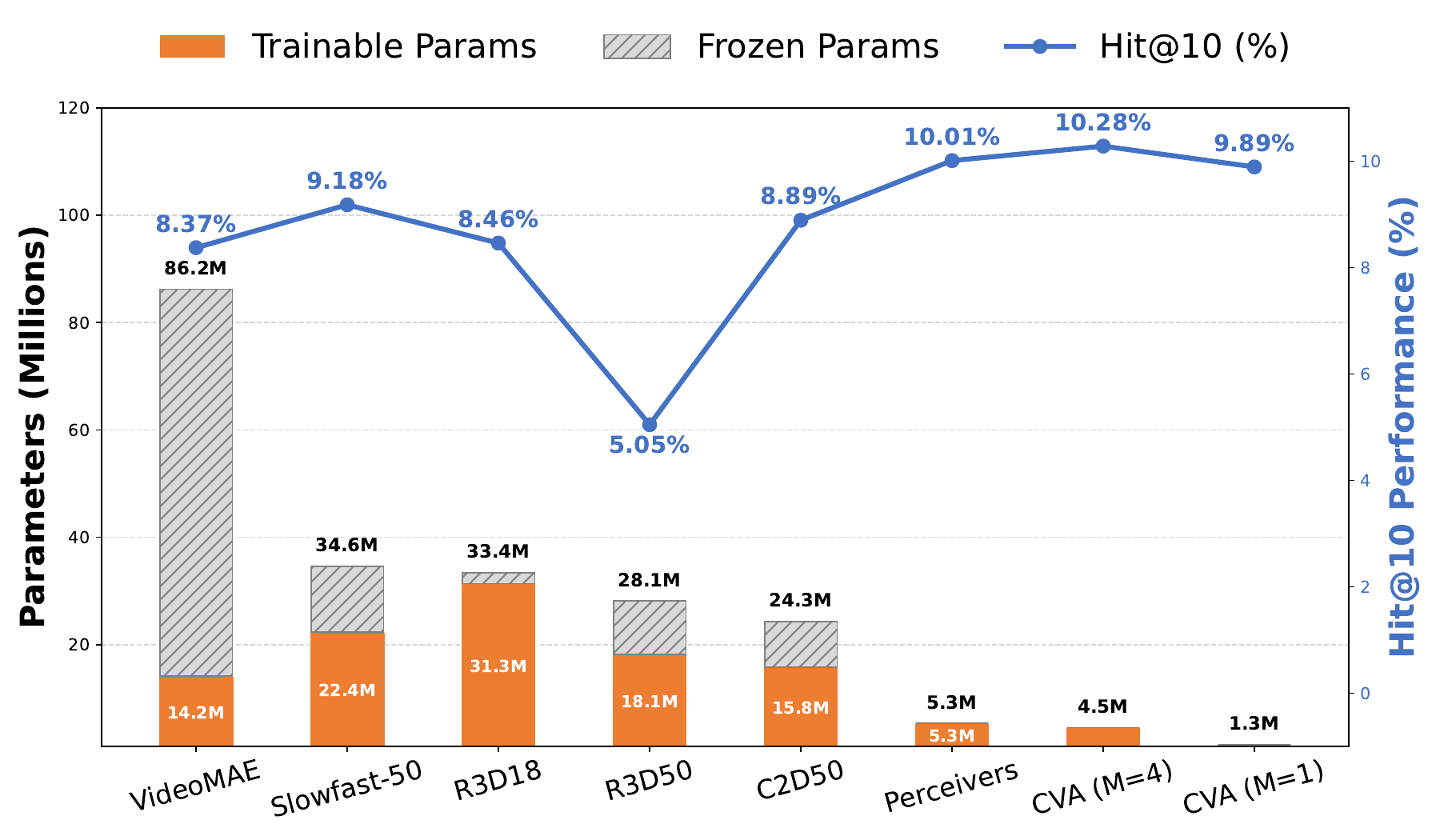}
    \caption{Comparison of parameters and performance. Video-backbone points follow the benchmark runs, while Perceiver IO and CVA use the controlled resampled-frame, batch-size-240, seed-42 setting. CVA ($M=4$) has only 4.5M fully trainable parameters and achieves a strong performance--efficiency trade-off.
    }
    \label{fig:models_params}
\end{figure}

Table~\ref{tab:experiments} and Table~\ref{tab:short_video_main} summarize the overall results on MicroLens and the Short-Video benchmark, respectively (Visualization can be seen in Figure~\ref{fig:models_params}).
Across all user encoders, our lightweight visual pipeline (\textit{frozen VFM} $\rightarrow$ \textit{CVA} $\rightarrow$ \textit{user encoder}) consistently improves recommendation quality while dramatically reducing computation (training time) and GPU memory, compared to traditional video backbones (For the non-visual content methods, see the Table~\ref{tab:non_video_only_experiments} in Appendix~\ref{sec: non-video-only}).

\textbf{MicroLens-100K.} 
Heavy video backbones (e.g., SlowFast-50~\cite{feichtenhofer2019slowfast}, VideoMAE~\cite{tong2022videomae}, ResNet-3D~\cite{du2021revisiting}) can achieve competitive accuracy but require prohibitively large training cost (e.g., more than 24k - 46k seconds and up to more than 100 GB GPU memory). In contrast, replacing expensive video encoding with our CVA yields both \emph{higher accuracy} and \emph{orders-of-magnitude efficiency}. 
For example, with SASRec, CVA ($M=4$) reaches \textbf{10.281} HR@10 and \textbf{14.723} HR@20, surpassing SlowFast-50 (9.179 HR@10) and VideoMAE (8.373 HR@10). This improvement comes with a \textbf{$\sim$33$\times$} reduction in training time (46513s $\rightarrow$ 1395s) and a \textbf{$\sim$250$\times$} reduction in peak memory (127.68GB $\rightarrow$ 0.51GB). With GRU4Rec, Perceiver IO obtains the highest accuracy, but exceeds CVA by only $0.031$ HR@10 while requiring $1.68\times$ the training time and $13.86\times$ the peak memory. Thus, CVA retains a substantially stronger efficiency--accuracy trade-off across user encoders. Full fine-tuning Qwen3-VL-2B modestly improves its LoRA counterpart by $+0.030/+0.062$ HR@10/NDCG@10 and $+0.064/+0.070$ HR@20/NDCG@20, but remains below the lightweight SASRec--CVA configuration while using substantially more memory.

\textbf{Short-Video.}
Although Short-Video does not provide ready-to-use user sequences and requires the preprocessing described in Appendix~\ref{sec: Data Preprocessing for the Short-Video Dataset}, CVA improves over Perceiver IO for most user encoders.
In particular, for NARM, CVA boosts HR@10 from 0.346 to \textbf{1.433}; for GRU4Rec and FMLPRec, CVA also brings consistent gains with comparable or lower training time and memory footprint.
These results confirm that our framework generalizes across different user-video datasets and recommendation architectures.

\begin{table}[t]
  \centering
  \small
  \setlength{\tabcolsep}{1.4mm}
  \begin{tabular}{lcccc}
    \toprule
    Visual Aggregator & HR@10 & NDCG@10 & HR@20 & NDCG@20\\
    \midrule
    Perceiver IO & 0.346 & 0.281 & 0.461 & 0.311\\
    CVA ($M=4$) & \textbf{1.433} & \textbf{0.875} & \textbf{1.713} & \textbf{0.947}\\
    \bottomrule
  \end{tabular}
  \caption{Cross-dataset evaluation with NARM on Short-Video. The complete user-encoder comparison is provided in Appendix~\ref{sec:short_video_appendix}.}
  \label{tab:short_video_main}
\end{table}

\begin{table*}[t]
  \begin{center}
    \begin{small}
      \begin{sc}
        \begin{tabular}{lcccccccr}
            \toprule
            \midrule
            User Encoder  & Visual Encoder & Sample & HR@10 & NDCG@10 & HR@20 & NDCG@20 & Time(s) & Mem. (GB)\\
            \midrule
            SASRec& Pooling& Mid 5   & 4.177 & 2.301& 5.928& 2.742& 730&0.64\\
            SASRec& Pooling& Resample & 4.914 (↑)& 2.576 (↑)& 7.083 (↑)& 3.123 (↑)& 738 & 0.64\\
            GRU4Rec& VideoMAE& Mid 5  & 8.221& 4.406& 12.145& 5.392& 39588&80.65\\
            GRU4Rec& VideoMAE& Resample & 8.607 (↑)& 4.724 (↑)& 12.706 (↑)& 5.754 (↑)& 40430&80.65\\
            SASRec& CVA (M=4)& Mid 5 & 9.591 & 5.177 & 13.922 & 6.268 & 1124 & 1.09\\
            SASRec& CVA (M=4)& Resample & 9.952 (↑)& 5.428 (↑)& 14.325 (↑)& 6.528 (↑)& 1131&1.09\\
            \midrule
            \bottomrule
        \end{tabular}
      \end{sc}
    \end{small}
  \end{center}
  \caption{Core resampling ablation on MicroLens-100K under the original batch-size-60, seed-42 protocol. These paired runs isolate frame selection and are not directly comparable to the batch-size-240 controlled aggregator rows in Table~\ref{tab:experiments}. The complete user-encoder and visual-encoder matrix is reported in Appendix~\ref{sec:resampling_analyses}.}
  \label{tab:Ablation_resampling}
\end{table*}
\subsection{Ablation Studies}
\textbf{Resampling.} Table~\ref{tab:Ablation_resampling} isolates frame selection across three substantially different visual architectures. For parameter-free mean pooling, resampling improves HR@10/NDCG@10 by $+0.737/+0.275$ and HR@20/NDCG@20 by $+1.155/+0.381$. With the heavy VideoMAE backbone, the corresponding gains are $+0.386/+0.318$ and $+0.561/+0.362$. CVA also improves by $+0.361/+0.251$ at cutoff 10 and $+0.403/+0.260$ at cutoff 20. Training time and peak memory remain comparable within each pair because resampling changes the selected offline inputs rather than the trainable online architecture. The consistent improvements for a fixed operator, an end-to-end video backbone, and the proposed compact aggregator show that resampling contributes independently of CVA. Dataset-level quality analysis is reported in Table~\ref{tab:desc_evaluation}, and the complete user-encoder and visual-encoder matrix is provided in Appendix~\ref{sec:resampling_analyses}.

Title corruption and missing-title results are reported in Appendix~\ref{sec:resampling_title_robustness}. Performance degrades under corrupted guidance, while remaining above the corresponding no-title setting in most configurations.

\textbf{Module of CVA.} Table~\ref{tab:controlled_aggregators} provides the primary controlled comparison, while the complete depth and attention-component sweep is reported in Appendix~\ref{sec:additional_cva_ablations}.

The parameter-matched MLP isolates whether CVA's gain comes from the attention parameterization or merely from additional capacity; Perceiver IO tests whether frame-to-latent cross-attention is required. CVA instead initializes refinement from a mean-projected, input-conditioned anchor. Its higher mean across all four metrics and lower variation than the matched MLP support this design without attributing pairwise temporal interaction to post-aggregation attention.


\section{Analysis}
\textbf{What does CVA refine?}
CVA is permutation-invariant over its frame inputs: masked mean pooling first forms a video-level semantic anchor, and attention is applied only after this aggregation. Thus, CVA does not claim pairwise temporal interaction between frames. The controlled three-seed comparison in Table~\ref{tab:controlled_aggregators} shows that post-aggregation attention achieves the highest mean on every metric, outperforming both an exactly parameter-matched MLP and Perceiver-style frame-to-latent cross-attention. Complete component, depth, long-tail, and VFM analyses are provided in Appendix~\ref{sec:additional_cva_ablations}; title robustness and frame-count analyses are provided in Appendices~\ref{sec:resampling_title_robustness} and~\ref{sec:resampling_frame_count}, respectively.

\textbf{Why is semantic resampling complementary to CVA?}
Semantic resampling improves the quality of the cached visual evidence before any trainable aggregation is applied, whereas CVA determines how that evidence is compressed for recommendation. The paired results in Table~\ref{tab:Ablation_resampling} show gains for parameter-free pooling, an end-to-end VideoMAE backbone, and CVA under otherwise identical settings. Therefore, the improvement cannot be explained solely by CVA's trainable capacity or latent refinement. Resampling suppresses semantically weak or redundant inputs offline, while CVA learns a compact task-oriented representation from the selected evidence online. Their benefits arise at different stages of the pipeline, making the two components complementary rather than mutually dependent.

\section{Conclusion}
We presented CVA, a lightweight content-driven module for micro-video sequential recommendation. 
CVA compresses frozen VFM frame embeddings into a compact video embedding through mean-based semantic consensus, input-conditioned latent expansion, and post-aggregation self-attention refinement.
Across two video-content recommendation benchmarks, CVA achieves a strong performance--efficiency trade-off, consistently improving performance while dramatically reducing training time and GPU memory compared with heavy video backbones.


\appendix
\section{Experimental Details}\label{sec:appendix_setting}

\paragraph{Implementation.}
All models are implemented in PyTorch. Unless otherwise stated, we use the AdamW optimizer with weight decay $1\mathrm{e}{-1}$.
Except for the controlled aggregator comparison, we set the random seed to 42 and report one run following the benchmark protocol. The controlled comparison uses seeds 1, 42, and 1000 as specified in the main paper; the individual runs are reported in Table~\ref{tab:controlled_aggregators_per_seed}.
All experiments are conducted on 4$\times$ NVIDIA L40S GPUs.

\paragraph{Batch-size protocols.}
Because institutional compute availability changed during the course of this study, the experiments were completed under two physical batch-size regimes. The controlled aggregator comparisons reported in the main paper use a physical batch size of 240, whereas several earlier and large-scale ablations retained the original physical batch size of 60. Changing the physical batch size also changes the number of optimizer updates and the optimization trajectory; consequently, entries with otherwise identical method labels and settings may exhibit small differences in their absolute results across tables. These entries should therefore be interpreted within their respective controlled protocols rather than as strict replications of one another. Within each protocol, the compared methods use the same data split, frame features, batch size, and evaluation procedure, and CVA retains an overall advantage over the relevant baselines in most controlled comparisons.

\paragraph{Data split.}
We adopt the standard leave-two-out protocol for sequential recommendation.
For each user sequence, we reserve the last interaction as the test target and the second last interaction as the validation target; the remaining prefix is used for training.
We truncate each sequence to the most recent $\texttt{max\_seq\_len}+3$ items before splitting.

\paragraph{Evaluation.}
We report Hit@K and NDCG@K with $K\in\{10,20\}$.
Negative samples are drawn from a popularity-based distribution $p(i)\propto \text{count}(i)^{\alpha}$.

\paragraph{Training.}
We train the user encoder and the visual projector (Perceiver IO / CVA / MLP) end-to-end, while keeping the VFM backbone frozen.
For all baselines using video backbones, we follow the benchmark setting and fine-tune only the last block.
For the Qwen3-VL experiments, validation-best CVA item features are frozen, while the Qwen sequential backbone and retrieval heads are optimized for recommendation. The LoRA variant updates rank-8 adapters, whereas Full FT updates all 1.722B parameters; both use batch size 60 and seed 42, and Full FT uses early stopping with patience 5.

\paragraph{Efficiency measurement.}
Training time (Time(s)) is measured as the wall-clock time of one full training run, including data loading and forward/backward passes.
GPU memory (Mem.(GB)) is the peak allocated memory during training, measured by \texttt{torch.cuda.max\_memory\_allocated}.

\begin{table*}[t]
    \centering
    \begin{tabular}{ccccccc}
    \toprule
        Dataset & User Encoder & Visual Encoder & Epoch & lr & batch size& GPU \\
    \midrule
        MicroLens & * & * & 30 & 1e-3 & 60 / 240$^\dagger$ & 4 L40S \\
        Short-Video & * & * & 60 & 1e-3 & 60 & 4 L40S \\
    \bottomrule
    \end{tabular}
    \caption{Default training settings. * refers to all methods; $^\dagger$ batch size 240 is used only for the controlled MLP--Perceiver IO--CVA comparisons in the main paper, while the remaining MicroLens ablations use batch size 60.}
    \label{tab:detailed setting}
\end{table*}
\subsection{Model Hyperparameters}
Full-block tuning was attempted in the benchmark~\cite{ni2025content} and found to degrade performance; we therefore follow the official recommended setting. Here we show the specific configuration of Perceiver IO and CVA (See in Table~\ref{tab:models config}).
\begin{table*}[t]
    \centering
    \begin{tabular}{cccccccc}
        \toprule
        Models & Dim. & Query Dim. & Depth (M) & Latent Form & Latent Dim. & \# of heads for cross. & \# of heads for self. \\
        \midrule
        Perceiver IO & 384 & 384 & 4 & 32 & 256 & 1 & 4 \\
        CVA (Ours) & 384 & -- & 4 & Projected anchor & 256 & -- & 4 \\
        \bottomrule
    \end{tabular}
    \caption{Models' Hyper-parameters}
    \label{tab:models config}
\end{table*}
\subsection{Baseline Fine-tuning Protocol}
Following the MicroLens benchmark~\cite{ni2025content}, visual-backbone baselines tune only the recommended final blocks. Table~\ref{tab:frozen_layer} shows that full fine-tuning substantially increases cost and can degrade accuracy.

\begin{table*}[t]
  \centering
  \small
  \begin{tabular}{llrrrrrrr}
    \toprule
    User Encoder & Visual Encoder & Frozen Layer & HR@10 & NDCG@10 & HR@20 & NDCG@20 & Time(s) & Mem.(GB)\\
    \midrule
    SASRec & SlowFast-50 & 270 & 7.298 & 3.936 & 10.550 & 4.756 & 20698 & 127.68\\
    SASRec & SlowFast-50 & 0 & 4.746 & 2.389 & 7.454 & 3.074 & 29211 & --\\
    SASRec & VideoMAE & 152 & 6.860 & 3.619 & 9.984 & 4.404 & 21072 & 80.75\\
    SASRec & VideoMAE & 0 & 0.052 & 0.021 & 0.082 & 0.028 & 30039 & --\\
    SASRec & R3D-18 & 30 & 6.780 & 3.556 & 10.092 & 4.389 & 23470 & 81.83\\
    SASRec & R3D-18 & 0 & 0.428 & 0.180 & 0.742 & 0.258 & 32197 & --\\
    \bottomrule
  \end{tabular}
  \caption{Effect of freezing different visual-encoder layers.}
  \label{tab:frozen_layer}
\end{table*}
\section{Additional MicroLens Results}\label{sec:additional_microlens_results}
\subsection{NextItNet Results}
Table~\ref{tab:nextitnet_main_appendix} reports the NextItNet results omitted from the main paper for space.
\begin{table*}[t]
  \centering
  \small
  \begin{tabular}{lccccccc}
    \toprule
    User Encoder & Visual Encoder & HR@10 & NDCG@10 & HR@20 & NDCG@20 & Time(s) & Mem. (GB)\\
    \midrule
    NextItNet & SlowFast-50 & 8.553 & 4.560 & 12.565 & 5.569 & 30952 & 127.70\\
    NextItNet & VideoMAE & 7.215 & 3.770 & 10.649 & 4.634 & 40682 & 80.76\\
    NextItNet & MLP & 6.408 & 3.362 & 9.662 & 4.180 & \textbf{805} & 0.74\\
    NextItNet & Perceiver IO & 8.108 & 4.316 & 11.867 & 5.262 & 1244 & 6.88\\
    NextItNet & CVA ($M=4$) & \textbf{9.046} & \textbf{4.797} & \textbf{13.321} & \textbf{5.873} & 1132 & \textbf{0.93}\\
    \bottomrule
  \end{tabular}
  \caption{Additional MicroLens-100K results with NextItNet.}
  \label{tab:nextitnet_main_appendix}
\end{table*}

\subsection{Non-visual Baselines}\label{sec: non-video-only}
This section presents the results of traditional and sequence-based recommender systems that rely solely on collaborative filtering and sequence IDs (non-visual methods), excluding any video content features. The purpose of these experiments is to establish a clear baseline and demonstrate the performance gain achieved by incorporating visual features.

Table~\ref{tab:non_video_only_experiments} summarizes the performance of several widely-used non-visual methods, including classic models like DSSM, LightGCN, NFM, and DeepFM, as well as sequence models like SASRec, GRU4Rec, and NEXTITNET when run on the MicroLens dataset without video input.

As shown in Table~\ref{tab:non_video_only_experiments}, the performance of the traditional non-visual methods is sub-optimal.

\begin{itemize}
    \item \textbf{Traditional Models}: Collaborative methods (DSSM, LightGCN) and factorization-based methods (NFM, DeepFM) achieve HR@10 values generally below 4.0 and NDCG@10 values below 2.0. This indicates that ID-only-based collaborative signals are insufficient to capture user preferences effectively on this dataset.
    \item \textbf{Sequence Models (ID-only)}: When sequence models (SASRec, GRU4Rec, NEXTITNET) are used without video features, their performance improves significantly over the traditional models, with SASRec achieving an HR@10 of 9.09 and NDCG@10 of 5.17.
\end{itemize}

Crucially, comparing the ID-only sequence models in Table~\ref{tab:non_video_only_experiments} with their visual counterparts in Table~\ref{tab:experiments} shows the value of the proposed video pipeline. SASRec obtains 9.09 HR@10 without video, 10.009 with Perceiver IO, and 10.281 with CVA ($M=4$). GRU4Rec improves from 7.82 without video to 9.821 with Perceiver IO and 9.790 with CVA.

In conclusion, these baseline experiments confirm that while sequence modeling itself is effective, the incorporation of video content and visual features is vital for achieving state-of-the-art results. The significant performance gap between the ID-only baselines and the best visual-enhanced models validates the necessity of our approach to effectively integrate video representations into the recommendation process.

\begin{table*}[t]
  \begin{center}
    \begin{small}
      \begin{sc}
        \begin{tabular}{lccccccc}
            \toprule
            \midrule
            \multicolumn{8}{c}{\textbf{MicroLens}} \\
            \midrule
            User Encoder  & Method & HR@10 & NDCG@10 & HR@20 & NDCG@20 & Time(s) & Mem. (GB)\\
            \midrule
            DSSM & No video  & 3.94 & 1.93 & 6.54 & 2.58& - & - \\
            LightGCN & No video   & 3.72 & 1.77 & 6.18 & 2.39 & - & - \\
            NFM & No video   & 3.13 & 1.59 & 4.80 & 2.01 & - & - \\
            DeepFM & No video   & 3.50 & 1.70 & 5.71 & 2.25 & - & - \\   
            SASRec & No video & 9.09 & 5.17 & 12.78 & 6.10 & - & - \\   
            GRU4Rec & No video & 7.82 & 4.23 & 11.47 & 5.15 & - & - \\   
            NEXTITNET & No video & 8.05 & 4.42 & 11.75 & 5.35 & - & - \\   
            MMGCL & Need Video & 2.61 & 1.22 & 4.25 & 1.64 & 4167 & 4.12\\  
            \bottomrule
        \end{tabular}
      \end{sc}
    \end{small}
  \end{center}
  \caption{Results of non-visual and other baselines on MicroLens-100K dataset.}
  \vskip -0.1in
  \label{tab:non_video_only_experiments}
\end{table*}

\section{Additional CVA Analyses}\label{sec:additional_cva_ablations}

\subsection{Per-Seed Controlled Aggregator Results}\label{sec:per_seed_aggregators}
Table~\ref{tab:controlled_aggregators_per_seed} reports the individual runs underlying the mean $\pm$ sample standard deviation results in Table~\ref{tab:controlled_aggregators}. All runs use SASRec, identical resampled DINOv3 features, physical batch size 240, and the same optimization protocol.

\begin{table*}[t]
  \centering
  \small
  \setlength{\tabcolsep}{3.2mm}
  \begin{tabular}{lcrrrr}
    \toprule
    Visual Aggregator & Seed & HR@10 & NDCG@10 & HR@20 & NDCG@20\\
    \midrule
    Perceiver IO & 1    & 10.226 & 5.573 & 14.633 & 6.684\\
    Perceiver IO & 42   & 10.009 & 5.420 & 14.331 & 6.510\\
    Perceiver IO & 1000 & 10.123 & 5.530 & 14.607 & 6.656\\
    \midrule
    Parameter-matched MLP & 1    & 9.508  & 5.148 & 13.714 & 6.204\\
    Parameter-matched MLP & 42   & 10.185 & 5.670 & 14.376 & 6.724\\
    Parameter-matched MLP & 1000 & 10.344 & 5.708 & 14.843 & 6.841\\
    \midrule
    CVA ($M=4$) & 1    & 10.066 & 5.549 & 14.440 & 6.653\\
    CVA ($M=4$) & 42   & 10.281 & 5.647 & 14.723 & 6.767\\
    CVA ($M=4$) & 1000 & 10.429 & 5.795 & 14.780 & 6.893\\
    \bottomrule
  \end{tabular}
  \caption{Per-seed results for the controlled SASRec aggregator comparison on MicroLens-100K. These runs correspond to the aggregate statistics reported in Table~\ref{tab:controlled_aggregators}.}
  \label{tab:controlled_aggregators_per_seed}
\end{table*}

\subsection{Component and Depth Study}
Table~\ref{tab:Ablation_CVA} reports the complete single-seed architectural sweep under the original batch-size-60, seed-42 protocol. The controlled batch-size-240, three-seed comparison in the main paper supersedes these runs for the primary CVA--Perceiver--MLP conclusion. The depth sweep shows a non-monotonic accuracy--efficiency trade-off, with $M=4$ selected as the default controlled configuration.

\begin{table*}[t]
  \centering
  \small
  \setlength{\tabcolsep}{2.4mm}
  \begin{tabular}{lrrrrrr}
    \toprule
    Visual Aggregator & \# Param. & HR@10 & NDCG@10 & HR@20 & NDCG@20 & Time(s) / Mem.(GB)\\
    \midrule
    Pooling & 0 & 4.914 & 2.576 & 7.083 & 3.123 & 738 / 0.64\\
    MLP & 1,772,928 & 8.047 & 4.335 & 11.784 & 5.276 & 791 / 0.80\\
    Perceiver IO & 5,321,344 & 9.508 & 5.122 & 13.808 & 6.205 & 1463 / 6.60\\
    \quad w/o first cross-attention & 4,637,568 & 9.741 & 5.249 & 14.279 & 6.393 & 1287 / 5.89\\
    \quad w/o all cross-attention & 4,504,960 & 9.750 & 5.269 & 14.081 & 6.358 & 1131 / 5.87\\
    \quad w/o all attention & 829,568 & 9.581 & 5.137 & 13.731 & 6.182 & 951 / 2.50\\
    CVA ($M=1$) & 1,348,864 & 9.607 & 5.191 & 13.927 & 6.279 & 899 / 0.62\\
    CVA ($M=2$) & 2,401,024 & 9.626 & 5.235 & 14.236 & 6.396 & 977 / 0.70\\
    CVA ($M=4$) & 4,504,960 & 9.952 & 5.428 & 14.325 & 6.528 & 1131 / 1.09\\
    CVA ($M=6$) & 6,609,664 & 9.800 & 5.239 & 14.079 & 6.316 & 1290 / 0.99\\
    CVA ($M=8$) & 8,713,984 & 10.007 & 5.381 & 14.394 & 6.486 & 1427 / 1.14\\
    CVA ($M=16$) & 21,339,520 & 9.905 & 5.379 & 14.408 & 6.513 & 2038 / 1.73\\
    \bottomrule
  \end{tabular}
  \caption{Complete single-seed component and refinement-depth study on MicroLens-100K with SASRec, using the original physical batch size 60 and seed 42. Here, MLP denotes the original two-layer GEGLU baseline applied after mean pooling and is not parameter-matched to CVA; the parameter-matched MLP in Table~\ref{tab:controlled_aggregators} is a separate controlled baseline.}
  \label{tab:Ablation_CVA}
\end{table*}

\subsection{Content-category Analysis}
\begin{figure*}[t]
    \centering
    \includegraphics[width=0.9\linewidth]{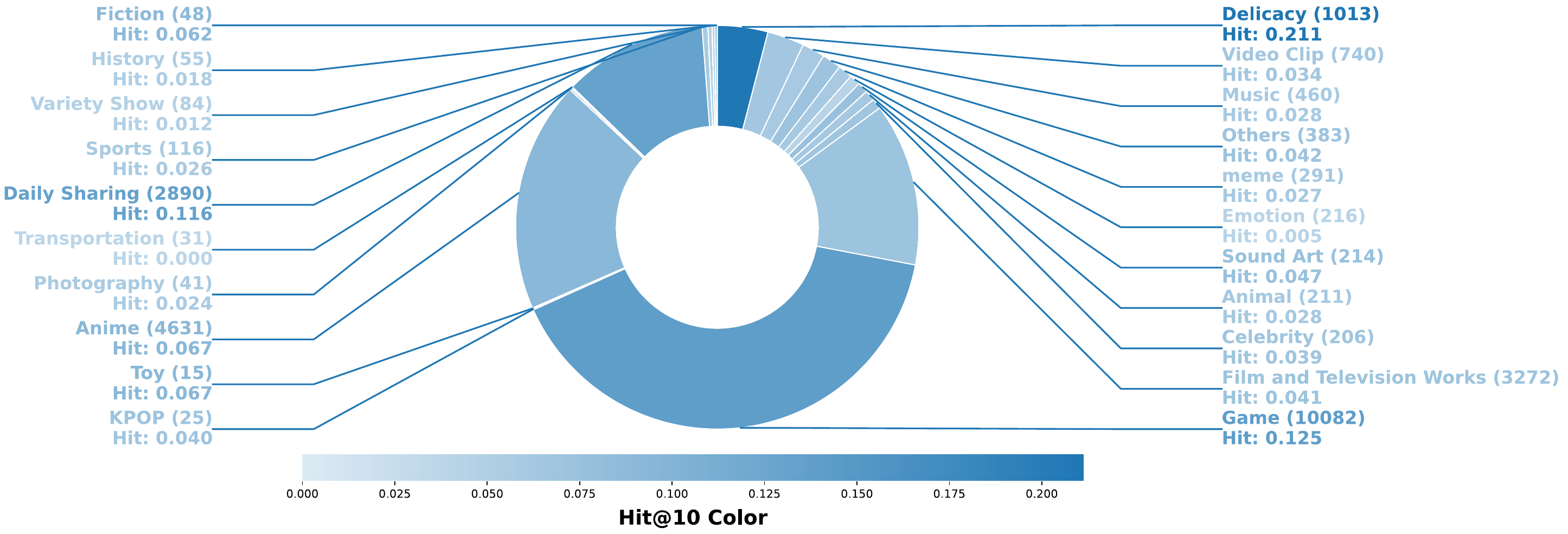}
    \caption{Per-tag statistics. Bars show Hit@10 and region sizes indicate the number of videos associated with each tag.}
    \label{fig:tags statistic}
\end{figure*}

Figure~\ref{fig:tags statistic} shows heterogeneous performance under the long-tail tag distribution; category frequency alone does not determine predictability.

\subsection{VFM Comparison}
CVA accepts frame-level embeddings from different frozen VFMs; Table~\ref{tab:different vfm} reports this comparison under the original batch-size-60, seed-42 protocol, with DINOv3 used as the default. Accordingly, its DINOv3 rows should not be read as repetitions of the batch-size-240 rows in Table~\ref{tab:experiments}.
\begin{table*}[t]
  \begin{center}
    \begin{small}
      \begin{sc}
        \begin{tabular}{lccccccc}
            \toprule
            \midrule
            User Encoder  & VFM & HR@10 & NDCG@10 & HR@20 & NDCG@20 & Time(s) & Mem. (GB)\\
            \midrule
            SASRec& Dinov3 & 9.952& 5.428& 14.325& 6.528& 1131&1.09\\
            GRU4Rec& Dinov3 & 9.467 & 5.078 & 13.644& 6.131& 1116&0.78\\
            NEXITNET& Dinov3 & 9.046 & 4.797 & 13.321& 5.873& 1132&0.89\\
            \midrule
            SASRec& Dinov2& 9.397& 5.044& 13.778& 6.147& 982&1.09\\
            GRU4Rec& Dinov2& 8.547& 4.579& 12.607& 5.599& 886&0.81\\
            NEXITNET& Dinov2& 8.152& 4.360& 12.044& 5.341& 989&0.93\\
            \midrule
            SASRec& Clip& 9.798& 5.343& 13.992& 6.397& 1150&1.09\\
            GRU4Rec& Clip& 9.166& 4.887& 13.166& 5.895& 979&0.97\\
            NEXITNET& Clip& 8.551& 4.703& 12.427& 5.677& 1140&1.15\\
            \midrule
            SASRec& Vit& 9.238& 4.959& 13.387& 6.005& 1136&2.19\\
            GRU4Rec& Vit& 8.554& 4.543& 12.646& 5.572& 933&1.77\\
            NEXITNET& Vit& 8.338& 4.459& 12.105& 5.407& 1145&1.96\\
            \bottomrule
        \end{tabular}
      \end{sc}
    \end{small}
  \end{center}
  \caption{Results of CVA with different frozen VFMs under the original physical batch size 60 and seed 42.}
  \vskip -0.1in
  \label{tab:different vfm}
\end{table*}

\subsection{Long-tail and Cold-start Analysis}
For the cold-start analysis in Table~\ref{tab:coldstart_experiments}, videos are divided into ten popularity groups according to interaction frequency.

\begin{table}[t]
  \begin{center}
    \begin{small}
    \setlength{\tabcolsep}{1mm} 
      \begin{sc}
        \begin{tabular}{lcccccc}
            \toprule
            \midrule
            User Encoder  & Group & Count & HR@10 & NDCG@10 & HR@20 & NDCG@20 \\
            \midrule
            \multirow{10}{*}{SASRec}& 1& 6235&  0.962&  0.469&  2.534&  0.859\\
            & 2& 5337&  2.005&  1.037&  4.141&  1.568\\
            & 3& 6396&  6.754&  3.282&  9.944&  4.079\\
            & 4& 6989&  4.493&  2.216&  7.197&  2.894\\
            & 5& 7756&  5.647&  2.683&  9.193&  3.573\\
            & 6& 9202&  8.868&  4.592&  12.486&  5.504\\
            & 7& 11179&  8.677&  4.203&  13.463&  5.41\\
            & 8& 12411&  9.516&  5.006&  14.27&  6.2\\
            & 9& 14736&  11.401&  6.249&  16.395&  7.506\\
            & 10& 19759&  18.989&  11.019&  25.35&  12.619\\
            \midrule
            \multirow{10}{*}{GRU4Rec}& 1& 6235&  0.497&  0.199&  1.508&  0.447\\
            & 2& 5337&  2.286&  0.932&  4.197&  1.413\\
            & 3& 6396&  3.361&  1.641&  5.66&  2.214\\
            & 4& 6989&  3.205&  1.563&  5.652&  2.175\\
            & 5& 7756&  4.551&  2.303&  7.671&  3.082\\
            & 6& 9202&  7.346&  3.599&  11.574&  4.665\\
            & 7& 11179&  9.491&  4.7&  14.026&  5.843\\
            & 8& 12411&  8.678&  4.584&  13.214&  5.726\\
            & 9& 14736&  10.24&  5.347&  15.174&  6.589\\
            & 10& 19759&  18.958&  10.9&  25.052&  12.441\\
            \midrule
            \multirow{10}{*}{NEXTITNET}& 1& 6235&  1.059&  0.58&  2.213&  0.869\\
            & 2& 5337&  1.705&  0.89&  3.822&  1.418\\
            & 3& 6396&  4.565&  2.162&  7.598&  2.93\\
            & 4& 6989&  3.348&  1.706&  5.595&  2.267\\
            & 5& 7756&  4.061&  2.19&  6.859&  2.897\\
            & 6& 9202&  7.238&  3.507&  10.433&  4.314\\
            & 7& 11179&  7.416&  3.875&  12.067&  5.048\\
            & 8& 12411&  7.808&  4.033&  11.997&  5.089\\
            & 9& 14736&  10.03&  5.364&  14.644&  6.526\\
            & 10& 19759&  18.26&  10.395&  24.622&  11.999\\
            \midrule
            \bottomrule
        \end{tabular}
      \end{sc}
    \end{small}
  \end{center}
  \caption{Experiments of Cold Start of CVA.}
  \label{tab:coldstart_experiments}
\end{table}

\section{Short-Video Dataset}\label{sec:short_video_appendix}
\subsection{Data Construction and Preprocessing}\label{sec: Data Preprocessing for the Short-Video Dataset}
Unlike MicroLens~\cite{ni2025content}, the sparsity of Short-Video Dataset~\cite{shang2025large} is extremely high (See in the Table~\ref{tab:Dataset Statistic on users} and Table~\ref{tab:Dataset Statistic on video}). 17,895 Short-Videos did not appear in users' metadata. And there is no direct user watching the history sequence data. Based on features such as user id (user ID(after hashing), each representing a real user of the platform), pid (video ID(after hashing), each representing a video collected from the platform), exposed time (Unix timestamp of the interaction), p date (date when the interaction happened), and p hour (the hour when the interaction happened) from the metadata, we constructed a data structure similar to MicroLens' to allow the model to read the data.

\begin{table*}[t]
    \centering
    \small
    \begin{tabular}{cccccccccccccc}
        \toprule
        \midrule
        Dataset & Valid Videos & N=1 & N=2 & N=3 & N=5 & N=10 & N=20 & Mean & Min. & Max. & 25\% & 50\% & 75\% \\
        \midrule
        MicroLens & 19738 & 2.34\% & 2.83\% & 3.16\% & 3.09\% & 2.41\%&1.60\% & 36.45 & 1 & 640 & 9 & 22 & 49\\
        Short-Video & 135666 & 27.36\% & 21.70\% & 14.56\%  & 6.05\% & 1.29\% &0.28\% & 4.73 & 1 & 798 & 1 & 3 & 5 \\
        \midrule
        \bottomrule
    \end{tabular}
    \caption{Datasets Statistic on Videos: Dense MicroLens and Sparse Short-Video}
    \label{tab:Dataset Statistic on video}
\end{table*}

A comparison between Table~\ref{tab:Dataset Statistic on video} and Table~\ref{tab:Dataset Statistic on users} highlights the core differences: MicroLens has a much larger number of users (100,000 vs. 6,893) but fewer videos (19,738 vs. 135,666). More critically, the average appearance frequency $N$ of videos in Short-Video is dramatically lower (4.75 vs. 36.44 in MicroLens). The sparsity is quantified by the fact that 27.36\% of videos in the Short-Video dataset appeared only once, compared to just 2.34\% in MicroLens. This signifies an extremely sparse User-Item Interaction Graph for the Short-Video data. This inherent high sparsity is the primary reason why the observed HR/NDCG results on the Short-Video Dataset are generally lower than those on MicroLens, though the relative performance against baseline methods remains comparable.

Given that our task is sequential recommendation—predicting the next video based on the history of played videos—it is essential to remove noisy data to improve the model's generalization ability. We performed the following cleaning steps:
\begin{itemize}
    \item \textbf{Video Pruning}: We removed videos that appeared only once or zero times. The raw metadata showed that 17,895 short-videos did not appear in any user's interaction history.
    \item \textbf{User Pruning}: We removed users with sequence lengths of 0 and 1.
\end{itemize}

Furthermore, to focus exclusively on users with sufficient behavioral information for effective sequential modeling, we further truncated the dataset by retaining only users whose sequence lengths were greater than or equal to 5. This processing resulted in a final set of 6,073 user sequence files used for training and evaluation. These steps ensure that the model is trained on quality sequences that reflect stable user behavior patterns, preventing interference from sparse, single-interaction noise.

\begin{table}[t]
    \centering
    \small
    \setlength{\tabcolsep}{1mm} 
    \begin{tabular}{cccccccc}
        \toprule
        \midrule
        Dataset & Valid Users & Mean & Min. & Max. & 25\% & 50\% & 75\% \\
        \midrule
        MicroLens & 100,000 &7.19 & 5 & 218 & 5 & 6 & 8\\
        Short-Video & 6893 & 93.63 & 1 & 2228 & 17 & 45 & 116\\
        \midrule
        \bottomrule
    \end{tabular}
    \caption{Dataset Statistic on Users' Sequence Length Statistics}
    \label{tab:Dataset Statistic on users}
\end{table}

\subsection{Complete Recommendation Results}

\begin{table*}[t]
  \begin{center}
    \setlength{\tabcolsep}{1mm}
    \begin{small}
      \begin{sc}
        \begin{tabular}{lccccccc}
            \toprule
            \midrule
            \multicolumn{8}{c}{\textbf{Short-Video}} \\
            \midrule
            User Encoder  & Visual Encoder & HR@10 & NDCG@10 & HR@20 & NDCG@20 & Time(s) & Mem. (GB)\\
            \midrule
            NARM& Perceiver IO& 0.346& 0.281& 0.461& 0.311& 603& 7.02\\
            NARM& CVA (M=4)& 1.433& 0.875& 1.713& 0.947& 576& 1.23\\
            \midrule
            NEXTITNET& Perceiver IO& 1.284& 0.846& 1.630& 0.933& 614& 7.13\\
            NEXTITNET& CVA (M=4)  & 1.416& 0.882& 1.713& 0.957& 589& 1.10\\
            \midrule
            GRU4Rec& Perceiver IO& 1.251& 0.837& 1.680& 0.946& 598& 6.69\\
            GRU4Rec& CVA (M=4)  & 1.466& 0.901& 1.762& 0.977& 577& 1.00\\
            \midrule
            FMLPRec& Perceiver IO& 1.400& 0.807& 1.696& 0.880& 614& 6.76\\
            FMLPRec& CVA (M=4)  & 1.433& 0.891& 1.877& 0.998& 581& 1.14\\
            \midrule
            BERT4Rec& Perceiver IO& 1.400& 0.882& 1.745& 0.968& 605& 7.06\\
            BERT4Rec& CVA (M=4)  & 1.416& 0.824& 1.828& 0.926& 575& 1.05\\
            \midrule
            SASRec& Perceiver IO& 1.416& 0.823& 1.828& 0.925& 605& 7.14\\       
            SASRec& CVA (M=4)  & 1.416& 0.884& 1.663& 0.947& 584& 1.14\\            
            \midrule
            \bottomrule
            
        \end{tabular}
      \end{sc}
    \end{small}
  \end{center}
  \caption{Results on Short-Video.}
  \label{tab:experiments_short_video}
\end{table*}
\section{Semantic Resampling Analyses}\label{sec:resampling_analyses}
\subsection{Full Resampling Ablation}\label{sec:full_resampling_ablation}
This section provides a complete ablation study on our semantic resampling strategy. All paired runs in this section use the original physical batch size 60 and seed 42; they isolate sampling effects and are separate from the batch-size-240 controlled aggregator comparison in the main paper.
We evaluate two sampling policies: the default MicroLens setting (\textit{Mid-5}) and our \textit{Resample} strategy, across three user encoders (SASRec, GRU4Rec, NextItNet) and multiple visual encoders, including heavy traditional video backbones (SlowFast-50, VideoMAE) and lightweight projection modules (Pooling, MLP, Perceiver, and CVA).

As shown in Table~\ref{tab:fully_ablation_resampling}, \textbf{Resampling consistently improves HR/NDCG across nearly all combinations}.
The gain is particularly evident for lightweight projection modules (Pooling/MLP/Perceiver/CVA), indicating that selecting more informative keyframes is crucial when the downstream model has limited capacity to recover semantics from noisy or redundant frames.
For heavy video backbones, resampling provides smaller but stable improvements, suggesting that strong temporal encoders can partially mitigate suboptimal frame selection but still benefit from higher-quality inputs.
Importantly, resampling does not change the model architecture and thus yields accuracy gains with comparable training cost.

\begin{table*}[t]
  \begin{center}
    \begin{small}
      \begin{sc}
        \begin{tabular}{lcccccccr}
            \toprule
            \midrule
            User Encoder  & Visual Encoder & Sample & HR@10 & NDCG@10 & HR@20 & NDCG@20 & Time(s) & Mem. (GB)\\
            \midrule
            
            \multirow{2}{*}{SASRec}& \multirow{2}{*}{Pooling}& Mid 5 & 4.177& 2.301& 5.928& 2.742& 730&0.64\\
            & & Resample & 4.914 (↑) & 2.567 (↑) & 7.083 (↑) & 3.123 (↑) & 738 & 0.64\\
            \midrule
            
            \multirow{2}{*}{SASRec}& \multirow{2}{*}{MLP}& Mid 5 & 7.563 & 4.108 & 11.255 & 5.035 & 801 & 0.80\\
            & & Resample & 8.047 (↑) & 4.335 (↑) & 11.784 (↑) & 5.276 (↑) & 791 & 0.80\\
            \midrule
            
            \multirow{2}{*}{SASRec}& \multirow{2}{*}{SlowFast-50}& Mid 5 & 9.179& 4.988& 13.072& 5.968& 46513&127.68\\
            & & Resample & 9.262 (↑)& 5.017 (↑)& 13.424 (↑)& 6.064 (↑)& 23617&127.68\\
            \midrule
            
            \multirow{2}{*}{SASRec}& \multirow{2}{*}{VideoMAE}& Mid 5 & 8.373& 4.456& 12.500& 5.495& 31571&80.75\\
            & & Resample & 8.504 (↑)& 4.585 (↑)& 12.604 (↑)& 5.618 (↑)& 24046&80.75\\
            \midrule
            
            \multirow{2}{*}{SASRec}& \multirow{2}{*}{Perceiver}& Mid 5 & 9.088& 4.898& 13.222& 5.939& 1230&6.87\\
            & & Resample & 9.508 (↑)& 5.122 (↑)& 13.808 (↑)& 6.205 (↑)& 1463&6.60\\
            \midrule
            
            \multirow{2}{*}{SASRec}& \multirow{2}{*}{CVA (M=4)}& Mid 5 & 9.591 & 5.177 & 13.922 & 6.268 & 1124 & 1.09\\
            & & Resample & 9.952 (↑)& 5.428 (↑)& 14.325 (↑)& 6.528 (↑)& 1131&1.09\\
            \midrule
            
            \multirow{2}{*}{GRU4Rec}& \multirow{2}{*}{Pooling}& Mid 5  & 3.771& 2.012& 5.471& 2.440& 611&0.51\\
            & & Resample & 4.574 (↑)& 2.289 (↑) & 6.773 (↑) & 2.845 (↑) & 611 & 0.51\\
            \midrule
            
            \multirow{2}{*}{GRU4Rec}& \multirow{2}{*}{MLP}& Mid 5 & 7.152 & 3.758 & 10.768 & 4.667 & 653 & 0.66\\
            & & Resample & 7.195 (↑)& 3.766 (↑) & 10.919 (↑) & 4.670 (↑) & 663 & 0.66 \\
            \midrule
            
            \multirow{2}{*}{GRU4Rec}& \multirow{2}{*}{SlowFast-50}& Mid 5  & 9.294& 5.021& 13.576& 6.097& 31044&127.59\\
            & & Resample & 9.487 (↑)& 5.203 (↑)& 13.648 (↑)& 6.251 (↑)& 27093&127.59\\
            \midrule
            
            \multirow{2}{*}{GRU4Rec}& \multirow{2}{*}{VideoMAE}& Mid 5  & 8.221& 4.406& 12.145& 5.392& 39588&80.65\\
            & & Resample & 8.607 (↑)& 4.724 (↑)& 12.706 (↑)& 5.754 (↑)& 40430&80.65\\
            \midrule
            
            \multirow{2}{*}{GRU4Rec}& \multirow{2}{*}{Perceiver}& Mid 5  & 9.052& 4.877& 13.144& 5.906& 1113&6.74\\
            & & Resample & 9.340 (↑)& 5.053 (↑)& 13.579 (↑)& 6.121 (↑)& 1119&6.51\\
            \midrule
            
            \multirow{2}{*}{GRU4Rec}& \multirow{2}{*}{CVA (M=4)}& Mid 5  & 9.311 & 5.041 & 13.562 & 6.112 & 1021 & 1.10\\
            & & Resample & 9.467 (↑)& 5.078 (↑)& 13.644 (↑)& 6.131 (↑)& 1116&1.10\\
            \midrule
            
            \multirow{2}{*}{NEXITNET}& \multirow{2}{*}{Pooling}& Mid 5  & 2.213 & 1.210 & 3.151 & 1.446 & 758 & 0.63\\
            & & Resample & 3.428 (↑) & 1.799 (↑) & 5.089 (↑) & 2.219 (↑) & 755 & 0.62\\
            \midrule
            
            \multirow{2}{*}{NEXITNET}& \multirow{2}{*}{MLP}& Mid 5 & 6.408 & 3.362 & 9.662 & 4.180 & 805 & 0.74\\
            & & Resample & 6.781 (↑) & 3.497 (↑) & 10.306 (↑) & 4.385 (↑) & 809 & 0.82 \\
            \midrule
            
            \multirow{2}{*}{NEXITNET}& \multirow{2}{*}{SlowFast-50}& Mid 5  & 8.553& 4.560& 12.565& 5.569& 30952&127.70\\
            & & Resample & 8.641 (↑)& 4.649 (↑)& 12.637 (↑)& 5.654 (↑)& 31062&127.70\\
            \midrule
            
            \multirow{2}{*}{NEXITNET}& \multirow{2}{*}{VideoMAE}& Mid 5  & 7.215& 3.770& 10.649& 4.634& 40682&80.76\\
            & & Resample & 7.335 (↑)& 3.829 (↑)& 10.726 (↑)& 4.681 (↑)& 39108&80.76\\
            \midrule
            
            \multirow{2}{*}{NEXITNET}& \multirow{2}{*}{Perceiver}& Mid 5  & 8.108 & 4.316& 11.867& 5.262& 1244&6.88\\
            & & Resample & 8.773 (↑)& 4.652 (↑)& 12.922 (↑)& 5.696 (↑)& 1485&6.60\\
            \midrule
            
            \multirow{2}{*}{NEXITNET}& \multirow{2}{*}{CVA (M=4)}& Mid 5 & 8.634 & 4.662 & 12.596 & 5.658 & 1124 & 1.11\\
            & & Resample & 9.046 (↑)& 4.797 (↑)& 13.321 (↑)& 5.873 (↑)& 1132&1.10\\
            
            \midrule
            \bottomrule
        \end{tabular}
      \end{sc}
    \end{small}
  \end{center}
  \caption{Full ablation results of the proposed resampling strategy on MicroLens under the original physical batch size 60 and seed 42.}
  \vskip -0.1in
  \label{tab:fully_ablation_resampling}
\end{table*}

\subsection{Robustness to Title Corruption}\label{sec:resampling_title_robustness}
Table~\ref{tab:titles_experiments} evaluates resampling with clean, missing, noisy, mismatched, and masked titles under the original physical batch size 60 and seed 42. These results diagnose the dependence of semantic resampling on metadata quality and are not part of the batch-size-240 controlled aggregator comparison.

\begin{table*}[t]
  \centering
  \small
  \setlength{\tabcolsep}{2.2mm}
  \begin{tabular}{llrrrrrr}
    \toprule
    User Encoder & Title Condition & HR@10 & NDCG@10 & HR@20 & NDCG@20 & Time(s) & Mem.(GB)\\
    \midrule
    GRU4Rec & Clean & 9.467 & 5.078 & 13.644 & 6.131 & 1116 & 0.78\\
    GRU4Rec & Without & 8.754 & 4.705 & 12.947 & 5.761 & 979 & 0.89\\
    GRU4Rec & Noise & 8.727 & 4.679 & 12.804 & 5.707 & 892 & 1.22\\
    GRU4Rec & Mismatched & 8.740 & 4.654 & 12.700 & 5.652 & 991 & 1.04\\
    GRU4Rec & Masked & 8.913 & 4.778 & 13.032 & 5.814 & 1103 & 1.36\\
    \midrule
    NextItNet & Clean & 9.046 & 4.797 & 13.321 & 5.873 & 1132 & 1.09\\
    NextItNet & Without & 8.181 & 4.361 & 11.951 & 5.309 & 1048 & 1.02\\
    NextItNet & Noise & 8.426 & 4.474 & 12.046 & 5.387 & 1114 & 1.07\\
    NextItNet & Mismatched & 8.406 & 4.498 & 12.389 & 5.499 & 1079 & 1.11\\
    NextItNet & Masked & 8.363 & 4.414 & 12.341 & 5.416 & 1001 & 1.21\\
    \midrule
    SASRec & Clean & 9.952 & 5.428 & 14.325 & 6.528 & 1131 & 0.84\\
    SASRec & Without & 9.704 & 5.178 & 14.110 & 6.287 & 1070 & 1.07\\
    SASRec & Noise & 9.596 & 5.148 & 14.110 & 6.286 & 992 & 0.93\\
    SASRec & Mismatched & 9.752 & 5.271 & 14.036 & 6.350 & 1039 & 1.02\\
    SASRec & Masked & 9.373 & 5.079 & 13.661 & 6.157 & 991 & 1.17\\
    \bottomrule
  \end{tabular}
  \caption{Robustness of title-guided resampling under different title conditions, using the original physical batch size 60 and seed 42.}
  \label{tab:titles_experiments}
\end{table*}

\subsection{Effect of Frame Count}\label{sec:resampling_frame_count}
In this section, we investigate the impact of the number of resampling video frames on the model's performance and computational efficiency. The experimental results are summarized in Table \ref{tab:frames_experiments}.

\begin{figure}[ht]
  \centering
  \includegraphics[width=0.95\linewidth]{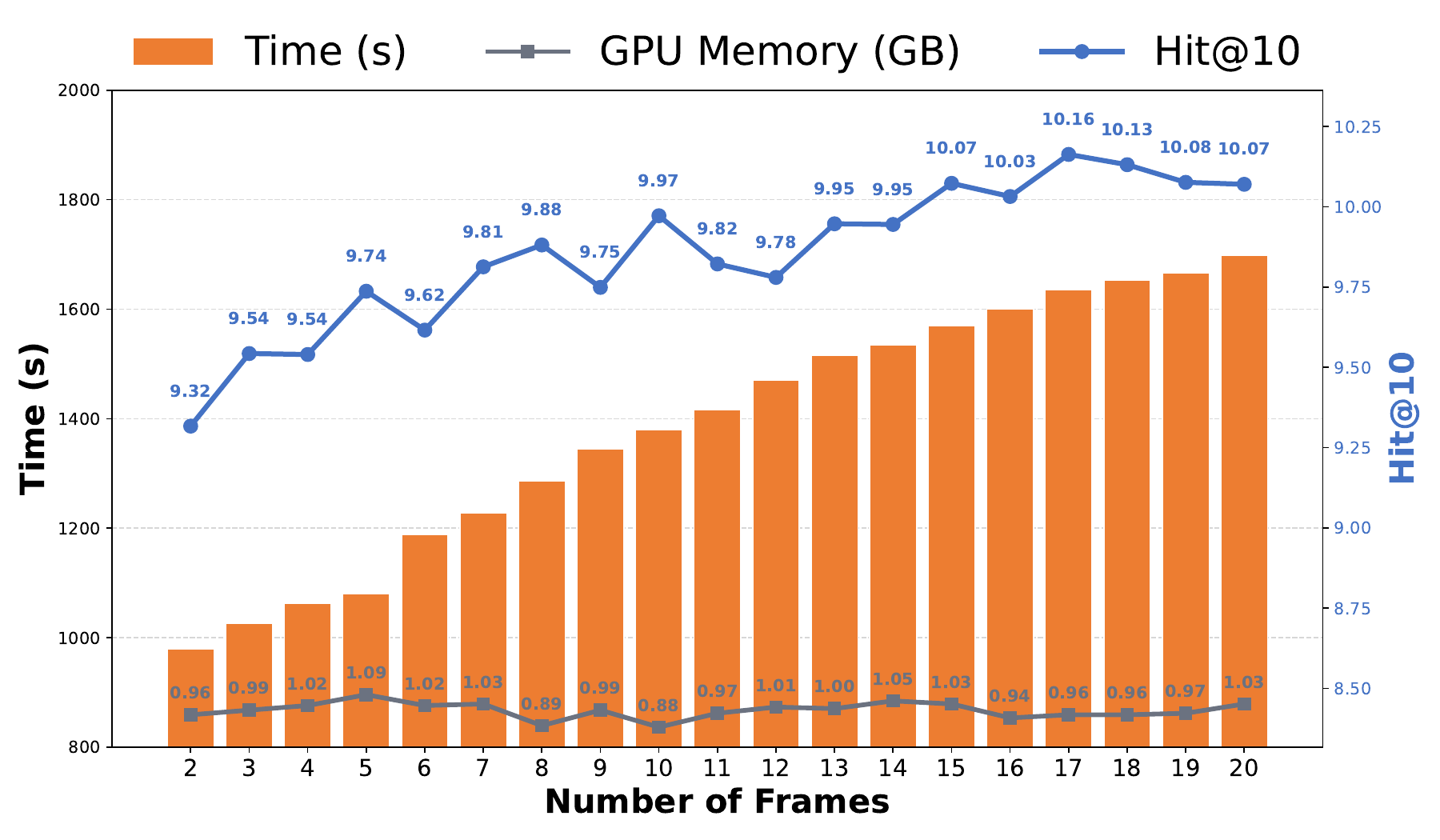}
  \caption{Recommendation performance with different numbers of input frames.}
  \label{fig:frames}
\end{figure}

First, increasing the number of sampled frames generally enhances the model's performance. As shown in the table, metrics such as HR@10 and NDCG@20 exhibit an overall upward trend as the frame count increases from 2 to 20. Specifically, the model achieves its peak performance (e.g., HR@20 of 14.663) at higher frame counts, suggesting that more frames provide broader visual and semantic coverage before mean aggregation.

Second, there is a clear trade-off between performance and training efficiency. While more frames improve accuracy, the training time per epoch increases consistently, rising from 979s for 2 frames to 1699s for 20 frames. This is primarily due to the increased overhead of data loading and feature extraction for the additional visual content.

Third, it is noteworthy that the GPU memory usage (Mem.) remains relatively stable despite the increase in frame count. This situation occurs because, in our CVA architecture, the frame-level embeddings undergo a pooling operation before being fed into the main model. Consequently, the input dimension for the subsequent layers remains constant regardless of the number of frames read, ensuring that the memory footprint does not scale linearly with the temporal resolution.

\begin{table*}[t]
  \begin{center}
    \begin{small}
      \begin{sc}
        \begin{tabular}{lcccccccr}
            \toprule
            \midrule
            User Encoder  & Visual Encoder & Frame & HR@10 & NDCG@10 & HR@20 & NDCG@20 & Time(s) & Mem. (GB)\\
            \midrule
            \multirow{19}{*}{SASRec}& \multirow{19}{*}{CVA (M=4)}& 2 &  9.317&  4.970&  13.617&  6.052& 979&0.96\\
            & & 3 &  9.543&  5.063&  13.852&  6.147& 1026&0.99\\
            & & 4 &  9.540&  5.077&  13.828&  6.156& 1063&1.02\\
            & & 5 &  9.737&  5.213&  14.114&  6.316& 1080&1.09\\
            & & 6 &  9.616&  5.166&  14.134&  6.300& 1188&1.02\\
            & & 7 &  9.813&  5.360&  14.337&  6.497& 1228&1.03\\
            & & 8 &  9.881&  5.276&  14.356&  6.404& 1287&0.89\\
            & & 9 &  9.749&  5.320&  14.172&  6.433& 1345&0.99\\
            & & 10 &  9.972&  5.405&  14.350&  6.505& 1380&0.88\\
            & & 11 &  9.822&  5.323&  14.341&  6.460& 1416&0.97\\
            & & 12 &  9.780&  5.309&  14.367&  6.462& 1471&1.01\\
            & & 13 &  9.947&  5.337&  14.394&  6.455& 1516&1.00\\
            & & 14 &  9.945&  5.332&  14.439&  6.466& 1535&1.05\\
            & & 15 &  10.073&  \textbf{5.525}&  \textbf{14.663}&  \textbf{6.684}& 1570&1.03\\
            & & 16 &  10.032&  5.327&  14.425&  6.432& 1601&0.94\\
            & & 17 &  \textbf{10.163}&  5.454&  14.591&  6.568& 1635&0.96\\
            & & 18 &  10.131&  5.399&  14.576&  6.518& 1653&0.96\\
            & & 19 &  10.076&  5.435&  14.630&  6.583& 1666&0.97\\
            & & 20 &  10.070&  5.492&  14.569&  6.625& 1699&1.03\\
            \midrule
            \bottomrule
        \end{tabular}
      \end{sc}
    \end{small}
  \end{center}
  \caption{Performance comparison of inputting different frames}
  \vskip -0.1in
  \label{tab:frames_experiments}
\end{table*}

\section{Limitations}

While our proposed CVA demonstrates a strong performance--efficiency trade-off, several limitations remain.

\textbf{Limited benchmark coverage.}
Our experiments are conducted on two publicly available video-content recommendation benchmarks, namely MicroLens and Short-Video. Although these datasets differ substantially in video density and user behavior distributions, they may not fully represent the diversity of industrial short-video platforms. Future work will evaluate CVA on larger-scale production datasets with more diverse content domains.

\textbf{Offline evaluation only.}
Our current study focuses on offline recommendation benchmarks, where training efficiency and ranking quality are measured under controlled settings. Real-world deployment introduces additional constraints such as online feature refresh, serving latency, cache consistency, and large-scale retrieval efficiency, which are beyond the scope of this work. We leave full online system evaluation to future research.

\textbf{Dependence on textual metadata during resampling.}
Our semantic resampling strategy optionally utilizes video titles to identify informative frames via CLIP similarity. While experiments with erroneous titles show certain robustness, the quality of textual metadata may vary across platforms, especially under missing, noisy, or clickbait-style titles. Exploring metadata-free or self-supervised frame selection remains an important direction.

\textbf{Limited motion modeling.}
Our framework relies on frozen visual foundation models that primarily encode semantic appearance rather than explicit temporal motion patterns. Although this design significantly improves efficiency, it may be less effective for motion-intensive videos where fine-grained temporal dynamics are critical. Incorporating lightweight motion-aware representations is another promising future direction.

\bibliography{aaai2027}


\end{document}